\documentclass[letterpaper, 10 pt, conference]{ieeeconf} 

\usepackage{times}
\usepackage{comment}
\usepackage{graphicx} 
\usepackage{xcolor}
\usepackage{amsmath}
\usepackage{hyperref}

\IEEEoverridecommandlockouts                              

\title{\LARGE \bf
Low cost, easily manufactured, highly flexible strain and touch sensitive fiber for robotics applications
}

\author{Christian Diaz Herrera,$^{1, 2}$ Srushti Raste,$^{1, 2}$ Simin Liu,$^{1, 2}$ Miles Modeste,$^{1, 2}$ Jiyang (Patton) Yin,$^{1, 2}$\\Katelyn McCall,$^{1, 2}$ Yuxing Jared Yao,$^{1, 2}$ Roopkamal Chahal,$^{1, 2}$ Simon Chidley,$^{1, 2}$ Trung Ha,$^{1, 2}$\\ T. David Westmoreland,$^{1, 3}$ Sonia Roberts$^{1, 2, \dagger}$
\thanks{This work was supported by NSF ERI award \#347708.}
\thanks{$^{1}$ Wesleyan University}%
\thanks{$^{2}$ Computer Science, Department of Mathematics and Computer Science}%
\thanks{$^{3}$ Chemistry Department}%
\thanks{$^\dagger$ {\tt\small sfroberts@wesleyan.edu}}%
}

\begin{document}

\maketitle
\thispagestyle{empty}
\pagestyle{empty}

\begin{abstract}

Existing stretch and touch sensors for robots are generally expensive with respect to at least one of material costs, required manufacturing equipment, or manufacturing time. We present and experimentally characterize a conductive fiber made using only inexpensive commercial off-the-shelf parts (conductive thread at \$0.07/ft, silicone tubing at \$0.94/ft) and tools (loop-style needle threader at \$2), which can be manufactured quickly (20 cm length in 2 minutes.) We demonstrate its use as a resistive strain sensor with three applications: Triggering a grasp in a pneumatically actuated assistive finger, sensing the pose of a pneumatically actuated robotic strap, and estimating the pose of a flexible solid. We also demonstrate that it can be used as a capacitive sensor with two applications: First, as a touch sensor which triggers a commercial robot arm to move, and second, as a near-field sensor enabling the robot arm to follow a moving hand. The capacitive sensors are knitted, showcasing the high flexibility of the fiber. We discuss methods for improving manufacturing scalability and their cost trade-offs. Finally, we demonstrate a method for repairing a cut fiber.

\end{abstract}


\section{Introduction}

\subsection{Background}

Strain (e.g., \cite{georgopoulou2021fabrication, gaochen2023}), touch \cite{xu2024cushsense}, and near-field \cite{erickson2019multidimensional} sensors can improve safety and comfort during human-robot interactions, and tactile robotic skins are a growing area of interest \cite{souriamjadi2020, roberts2021soft, si2023robotsweater, xu2024cushsense}. %
However, many sensors use materials that are either expensive or difficult for people outside of research labs to obtain or manufacture, such as liquid metals \cite{souri2020wearable}, conductive polymers \cite{souri2020wearable}, conductive elastomers \cite{souri2020wearable}, and carbon derivatives such as nanotubes \cite{yangperception2024}, graphene, and carbon black nanoparticles \cite{yan2018flexible}. %

Design choices must balance desirable material properties (flexibility, elasticity), electrical properties (low noise, consistent readings), and manufacturing considerations like cost, accessibility of materials, accessibility of manufacturing equipment, and complexity of manufacturing procedure. %
For example, resistive strain and touch sensing requires only simple circuitry \cite{si2023robotsweater, roberts2021soft}, but is notoriously more noisy than capacitive sensing (e.g., \cite{xu2024cushsense}). %
However, circuits which measure changes in capacitance are much more complex and expensive to manufacture than voltage dividers \cite{xu2024cushsense, hanson2024controlling}. %

Most of the sensors relevant for robotics applications are developed for large-scale manufacturability at the cost of accessibility to DIY makers \cite{souriamjadi2020}. %
People outside of the large-scale manufacturing space or the well funded research laboratory may have different sensor design needs than those with easy access to resources. %
For example, those interested in improving their own prosthetics \cite{cossovich2024co}, building tools to serve their community, or working in educational settings or in under-resourced research groups may not be able to make use of currently available strain sensors which require expensive equipment or materials to manufacture \cite{hofmann2016clinical}. %
However, DIY makers are creative, and fully capable of making components on their own when accessible manufacturing methods are available \cite{jones2024hand}. %
We target this group with our current work. %

We present a sensor fiber made from inexpensive off-the-shelf materials, requiring no complex equipment to create. %
With a highly flexible, elastic, fiber-based form factor, the sensor can be used in a large variety of robotics applications, some of which we demonstrate in Sections \ref{sec:resistive_applications} and \ref{sec:capacitive_applications}. %
This sensor fiber can be used for both resistive strain sensing applications and capacitive touch or near-field sensing applications, making it suitable for a wide range of roboticists. %

\subsection{Contributions}
\begin{itemize}
    \item Manufacturing process of sensor fiber
    \item Characterization of fibers as resistive sensors
    \item Characterization as capacitive sensors
    \item Demonstration of repairability and behavior of fiber under adverse conditions
    \item Feasibility tests of scalable manufacturing adaptations
    \item Demonstrations in multiple robotics applications
\end{itemize}

\begin{figure}
    \centering
    \vspace{10px}
    \includegraphics[width=\linewidth]{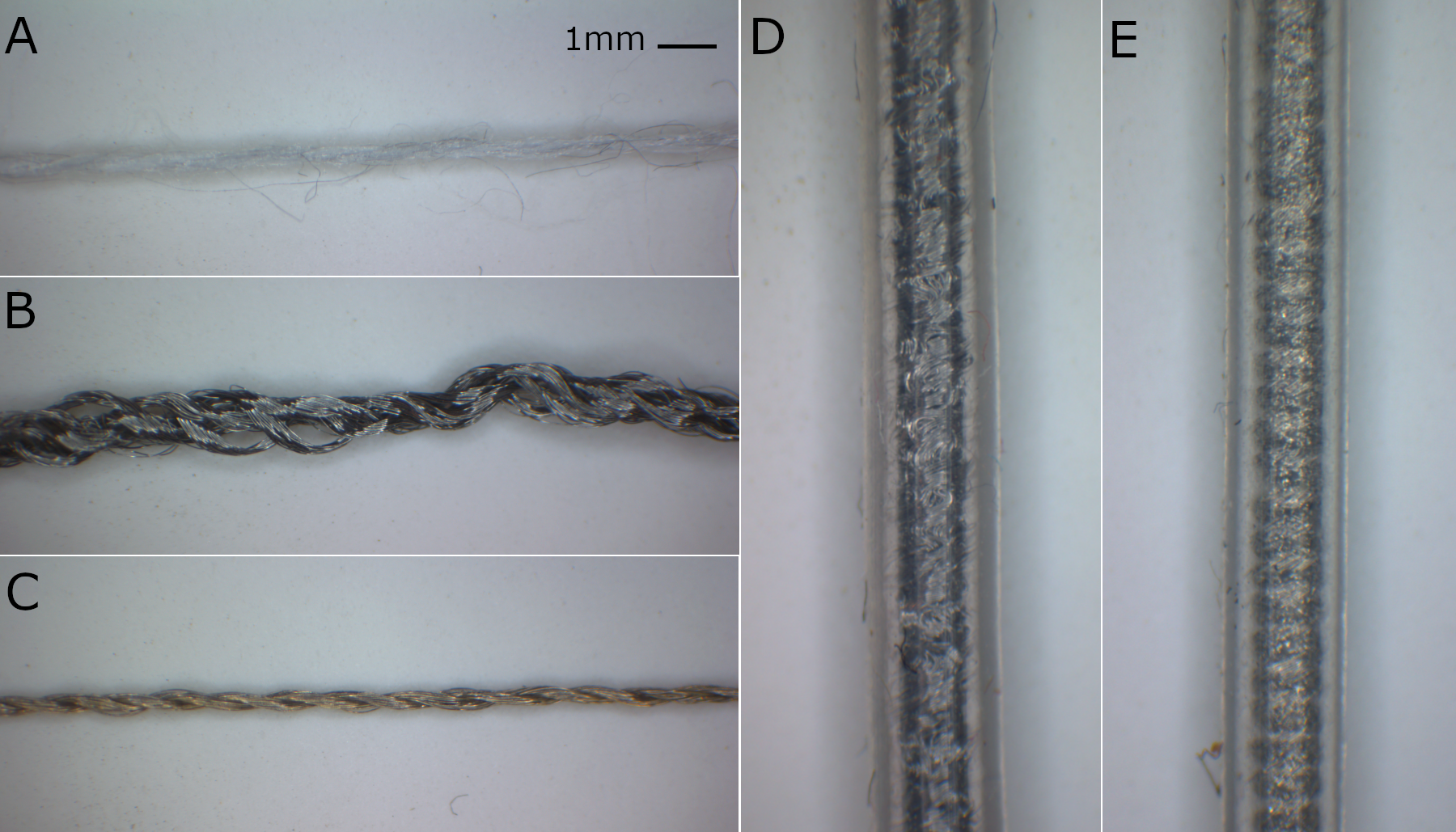}
    \caption{Three samples of conductive thread. A) Composite thread spun from nylon and stainless steel staple fibers. B) First electroplated thread. C) Second electroplated thread. D) Sensor fiber made from first electroplated thread. E) Sensor fiber made from second electroplated thread.}
    \label{fig:material_selection_thread}
\end{figure}

\section{Materials and manufacturing process}

The fibers characterized in Sections \ref{sec:resistive_experiments} and \ref{sec:capacitive_experiments}, and demonstrated in Sections \ref{sec:resistive_applications} and \ref{sec:capacitive_applications}, are made from two off-the-shelf parts: Electroplated thread and silicone tubing. %
We also tested some alternative materials to determine what conditions must be met for material substitutions to be successful (Section \ref{sec:materialselection}). %
To make the fibers, we used a loop-style needle threader (Section \ref{sec:manufacturing}). %

\subsection{Material selection and comparison}
\label{sec:materialselection}

\subsubsection{Conductive thread}
We used an electroplated thread for our sensors (Kitronik Ltd, manufacturer part number 2744) made by electroplating individual acrylic staple fibers and then spinning them together rather than plating an already spun thread. %
We tested two additional conductive threads for suitability: Stainless steel thread (spun from 100\% stainless steel staple fibers; SparkFun part number DEV-11791) and composite conductive thread (80\% acrylic, 20\% stainless steel staple fibers spun together; purchased from Yeoman's Yarns). %
The composite thread had approximately 1000 times the resistance as the electroplated thread ($40~\Omega$ vs 40  40 k$\Omega$ for a 20 cm sample). %
The stainless steel thread and electroplated threads had similar conductivity. %

We purchased a second sample of the electroplated thread from the same company to investigate the consistency of the product. %
The thread from the two samples differed qualitatively: The first sample was darker in color, softer, and more inclined to split. %
We called Kitronik to ask about the difference in the fibers and were told that the company sources from multiple mills which have slight differences in their manufacturing process, including materials used to coat the threads as they pass through the equipment, and that it is not unexpected to see some variation in the final product. %
We tested both Kitronik threads in the initial characterizations. %

We used a 5.5 digit multimeter (Keithley 2110) to compare variation in resistance over 10 minutes for five 20-cm samples each of the two electroplated threads, the composite thread, and the stainless steel thread. %
The first electroplated thread had an average drift in resistance of 0.3\% and the second had an average drift of 0.4\%, compared with 0.3\% for the stainless steel thread and 9.6\% for the composite thread. %
We rejected the composite thread from further consideration due to this extremely high, and unexplained, drift. %

We made sensor fibers from both electroplated threads and the stainless steel thread to test drift in resistance at rest over 10 minutes. %
All three fibers were relatively quiet, with drift of less than 0.4\%. %
Qualitatively, we also noticed that because the stainless steel thread was stiffer than the electroplated thread, it created lumps in the finished fiber. %
Based on these comparison experiments, we consider the stainless steel thread a viable alternative option for the capacitive sensing applications, but suggest using a softer thread for resistive strain sensing. %
We used the first electroplated thread for the resistive strain sensing characterizations in Section \ref{sec:resistive_experiments}, and the second electroplated thread for the capacitive pressure sensing characterizations in Section \ref{sec:capacitive_experiments}. %

\subsubsection{Silicone tubing}

We made sample sensors using silicone, PTFE, and latex tubing. %
We preferred silicone over other tubing materials due to silicone's high flexibility and elasticity, ubiquitous availability in a large range of sizes, and hypoallergenic properties. %
For the characterizations and demonstrations reported in this paper, we used the smallest silicone tubing available from McMaster-Carr (durometer 70A), with an outer diameter of 1/16" (1.6 mm) and an inner diameter of 1/32" (0.79 mm). %
However, we were able to successfully create sensor fibers using silicone tubing with as small an outer diameter as 0.5 mm (inner diameter 0.25mm; purchased from Avidmax) with the method described in Section \ref{sec:manufacturing}. %
For this smaller fiber, we used a \#13 beading needle and the second (stiffer) eletroplated thread purchase, which was thinner than the first (Figure \ref{fig:material_selection_thread}). %

\subsection{Manufacturing process}
\label{sec:manufacturing}

To manufacture the fibers, the only equipment required is a thin, dull needle or needle-like tool (hereafter the ``needle'') which can guide the electroplated thread through the silicone tubing without piercing the side walls. %
We were able to successfully make fibers using a loop-style needle threader (Dritz, part number 252), a beading needle (Beadsmith size 10-15 depending on tubing size), and a collapsible beading needle (FIVEIZERO). %
A detailed procedure follows. 

\begin{enumerate}
    \item Cut silicone cord to the desired length
    \item Thread electroplated thread through needle
    \item Feed needle into silicone tubing
    \item Stretch silicone tubing, pulling in more electroplated thread, and then release
    \item Continue moving the needle through the silicone tubing, stretching and releasing the silicone tubing to pull in more electroplated thread
\end{enumerate}

\noindent For a 20 cm length, the manufacturing process takes under 2 minutes. See the supplementary video for a demonstration. 

\section{Resistive sensor characterizations}
\label{sec:resistive_experiments}

\subsection{Principle of operation for resistive sensing}
\label{sec:principle_resistive}

\begin{figure}
    \centering
    \vspace{10px}
    \includegraphics[width=\columnwidth]{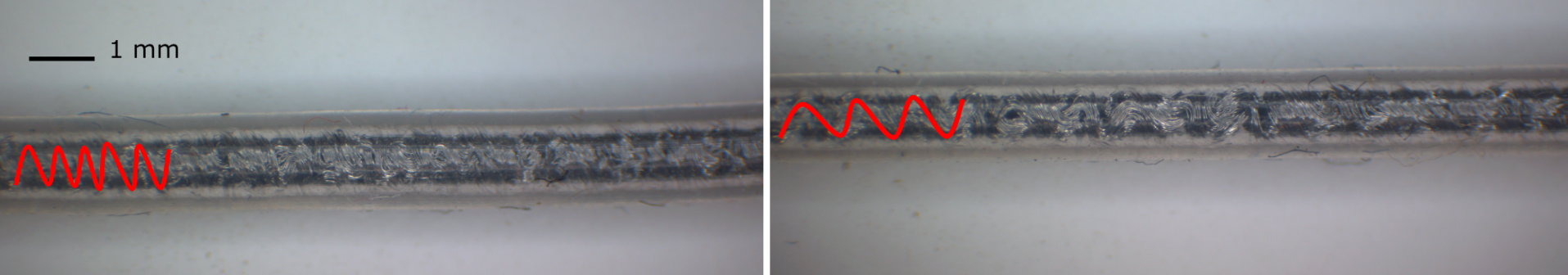}
    \caption{A sensor fiber photographed at 10x magnification, at rest (left) and stretched to 30\% strain (right). Stretching the fiber unfolds the electroplated thread inside the tubing. The red line shows how the electroplated thread is folded inside the tube. }
    \label{fig:fiber_stretch}
\end{figure}

The process of repeatedly stretching the tubing to draw in electroplated thread stuffs the silicone tubing with the electroplated thread, which folds into an accordion-like structure (Figure \ref{fig:fiber_stretch}). %
This creates a thick but extremely flexible ``wire.''. %

As the sensor fiber is stretched, the electroplated thread unfolds (Figure \ref{fig:fiber_stretch}). %
This increases the length of the shortest path that an electron must travel in order to traverse the length of the sensor fiber. %
Functionally, this is equivalent to increasing the length of the ``wire'' formed by the folded electroplated thread inside of the tubing and decreasing its cross-sectional area, leading to an increase in resistance. %

\subsection{Resistive strain sensing experiments}
\label{sec:exp_strain}

\begin{figure}
    \centering
    \vspace{10px}
    {\tiny Variation across five fibers \hspace{60px} Five stretch-relaxation cycles}
    \includegraphics[height=93px]{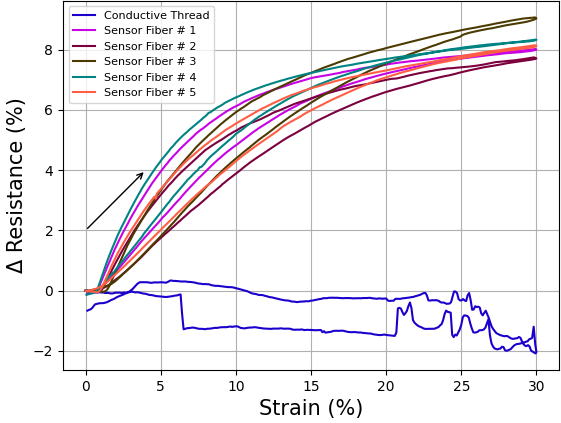}
    \includegraphics[height=93px]{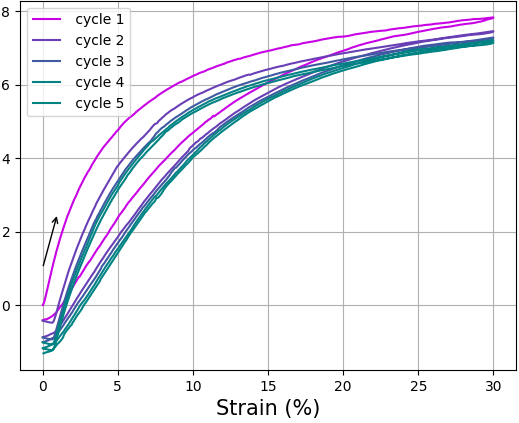}
    \caption{Resistive stretch sensor characterization. We stretched five sensor fibers, and one piece of electroplated thread, to 30\% strain five times. The arrow indicates the start of the data trace in time. (Left) The last stretch-relaxation cycle for all five sensor fibers and electroplated thread. Note that the electroplated thread does not change its resistance when stretched. (Right) All stretch-relaxation cycles for one sensor fiber to show drift. }
    \label{fig:stretch_cycles_short}
\end{figure}

\subsubsection{Predicting resistance at rest from length}
We measured the resistance of the sensor fibers at rest on a bench using a Keithley 2110 5.5 digit multimeter in lengths ranging from 2.1 to 45.3 cm, taking the average resistance over a period of 2 minutes at a capture rate of 50 Hz. %
These experiments were performed at room temperature in the lab but the temperature was not explicitly controlled. %
We then predicted rest resistance from length using a least squares linear regression. %
After measuring the resistance in the test leads as less than 0.1 $\Omega$, we assumed an intercept of 0 a linear relationship $R = c \cdot l$ for $R$ the resistance of the fiber in ohms and $l$ the length of the fiber in centimeters. We performed a least-squares linear fit to find the coefficient $c = 2.4583$, with coefficient of determination $R^2 = 0.9949$. %

\subsubsection{Effect of temperature on resistance}
Temperature is known to affect resistance. %
We placed one 42 cm sensor in 11 water baths with temperatures ranging from 13 to 66 C and predicted resistance at rest from temperature: $\Omega = 0.065326 \cdot \text{C} + 101.00$, $R^2 = 99\%$. %

\subsubsection{Characterizing change in resistance with stretch}
\label{sec:res_exp}
We used an Ametek Chatillon CS225 universal testing machine and 2-wire resistance readings from a Keithley 2110 5.5 digit multimeter to characterize the change in resistance of the fibers as they were stretched to 30\% strain. %
We recorded force and displacement readings using Ametek's CS2 series software, and resistance measurements using a custom Python script. %
Data from the universal testing machine and the multimeter were collected on the same computer and synced using timestamps. %
Based on the difference of time measurement precision between the two devices and the speed at which the tests were performed, the greatest possible physical offset is 0.16667 mm. 

Data from five stretch-relaxation cycles for five stretch sensitive fibers of approximately 20 cm are presented in Figure \ref{fig:stretch_cycles_short}. %
All fibers were stretched at a rate of 10 mm/min to 30\% strain. %
For $R_i$ the resistance at timestep $i$, and $R_0$ the first resistance measurement recorded, we calculated the percent change in resistance as $(R_i-R_0)\cdot (R_0)^{-1}$.
The relationship between strain and percent change in resistance has two linear regions connected by an elbow. %
All five sensors experienced approximately an 8\% change in resistance with stretch to 30\% strain. %
From these experiments, we determined that the gauge factor of our sensor is 0.28 at 30\% strain (standard deviation: 0.0240) and 0.6 at 10\% strain. %
The hysteresis is 19.5\% at 30\% strain (standard deviation: 0.0141), with a signal-to-noise ratio, calculated using a fast fourier transform, of 102.52. %

We also stretched one length of electroplated thread to 30\% strain by itself (Figure \ref{fig:stretch_cycles_short}). %
The electroplated thread did not change its resistance with strain, suggesting that the strain sensitivity is due to the structure of the fibers, not from the material properties of the electroplated thread. %

\begin{figure}
\vspace{10px}
    \centering
    \begin{minipage}{0.71\linewidth}
    \centering
    {\tiny \qquad Near-field and touch capacitive sensing}
    \includegraphics[width=\linewidth]{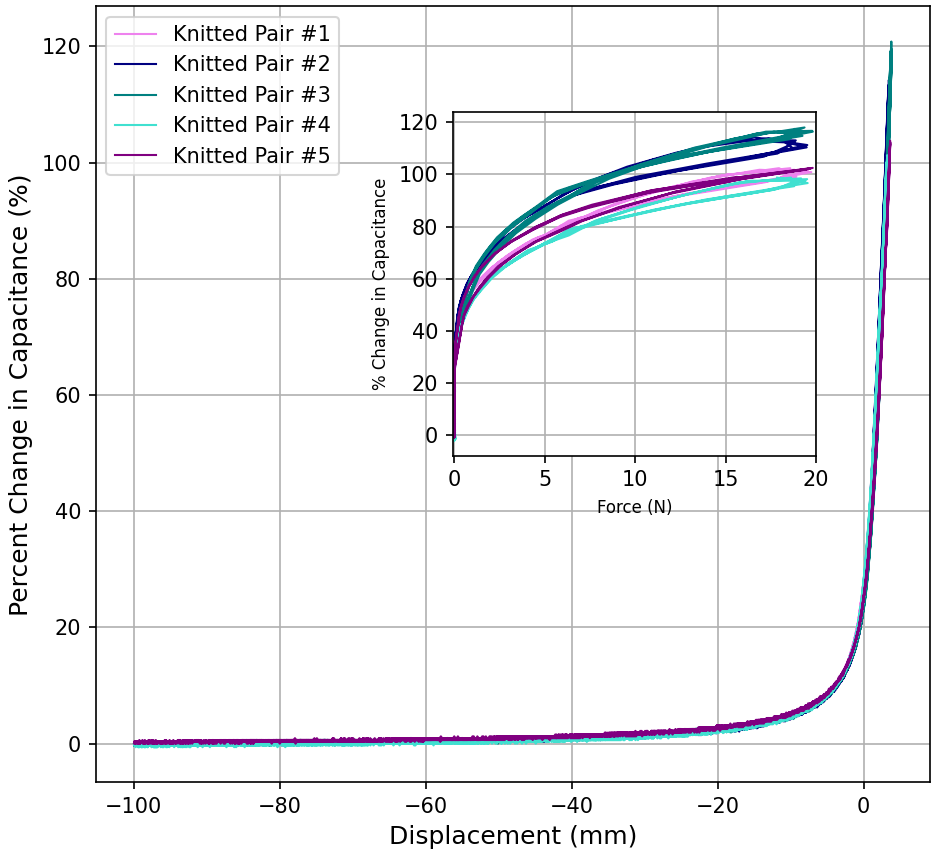}
    \end{minipage}
    \begin{minipage}{0.27\linewidth}
    \raggedleft
    \includegraphics[width=\linewidth]{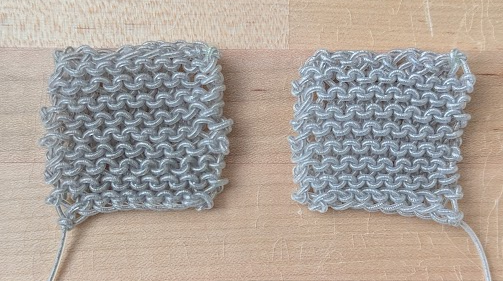}
    \\[2px]
    \includegraphics[width=\linewidth]{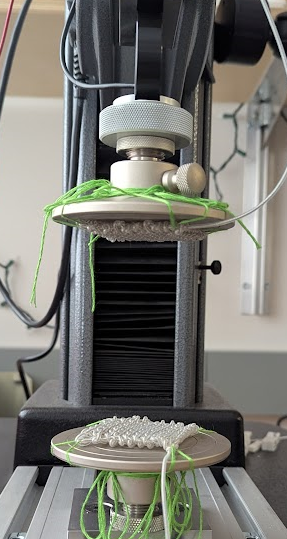}
    \end{minipage}
    \caption{(Left) Percent change in capacitance from baseline (measured when the knitted electrodes are 100 mm apart) against displacement. Positive mm values indicate compression of the two knitted electrodes against each other, negative values indicate distance between the electrodes. Inset shows the percent change in capacitance from baseline against force. (Right) Two knitted electrode patches side by side, and then paired to make a capacitive sensor.}
    \label{fig:capacitive_displacement}
\end{figure}

\section{Capacitive sensor characterizations}
\label{sec:capacitive_experiments}

\subsection{Principle of operation for capacitive sensing}
\label{sec:principle_capacitive}

A capacitor is built from two electrically conductive components separated by a dielectric. %
When the electrically conductive components are closer together, capacitance is higher, and when they are further away, capacitance is lower. %
The electroplated thread in our sensor fibers is conductive, and silicone is a dielectric. 
Because capacitance can be affected by the proximity of any conductive materials (including human bodies), capacitive touch sensors are generally made from two conductive components that are placed close together and then shielded. %
Near-field capacitive sensors can be made from one piece of conductive material, with air serving as the dielectric. %
In this case, both air and the silicone tubing surrounding the electroplated thread act as dielectrics. 

\subsection{Capacitive pressure and near-field sensing experiments}

To create capacitive pressure and near-field sensors, we hand knitted five pairs of two 5 cm x 5 cm squares, with each square being knit from 10' of sensor fiber using garter stitch. 
To hand knit garter stitch, knit every row. 
To machine knit garter stitch, flip the knitting or switch between opposing needle beds every row. %
This stitch was selected because it lays flat, whereas the plain knitting stitch on a knitting machine (stockinette in hand knitting terms) curls \cite{abel2013hierarchical}. %

We attached one square in each pair to a compression platen by tying the corners to the compression platen's mounting bracket. %
We used an Ametek Chatillon CS225 universal testing machine to displace the compression platen-mounted patches of knitted sensor fiber and measured capacitance with a Keithley 2110 5.5 digit multimeter. %
Displacement and force data were recorded using Ametek's CS2 series software, and capacitance measurements were recorded using a custom Python script. %

The experimental procedure was as follows. %
We aligned the squares on the platens visually, corner to corner, when loading the platens onto the universal testing machine. %
The top platen was moved by hand to a distance of approximately 1 cm above the lower platen, and load was zeroed. %
The universal testing machine then lowered the top platen at a rate of 10 mm/min until a reading of 0.03 N was reached, indicating that the knitted squares were just barely touching. %
The position was then zeroed and the top platen moved up to a displacement of -100 mm (negative numbers indicating distance away from the bottom platen) at a rate of 100 mm/min. %
The machine then performed five cycles of compression to 20 N and displacement 100 mm away from the bottom platen (Figure \ref{fig:capacitive_displacement}) to characterize the near-field and pressure-sensing capabilities of knitted patches of the sensor fiber. %

\section{Sensor degradation and repair}
\label{sec:repair}

One strength of the proposed sensor fibers is their resilience to adverse conditions and repairability. %
In this section we discuss the effects of long-term wear and extreme temperatures, and discuss a method for repairing a cut fiber. %

\subsection{Repairing cut fibers} 
\begin{figure}
    \centering
    \vspace{10px}
    \begin{minipage}{0.34\linewidth}
    \includegraphics[width=\linewidth]{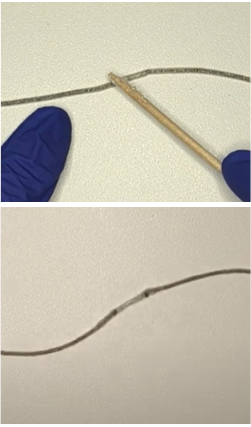}
    \end{minipage}
    \begin{minipage}{0.64\linewidth}
    \centering
    {\tiny \qquad Repaired sensor fiber}
    \includegraphics[width=\linewidth]{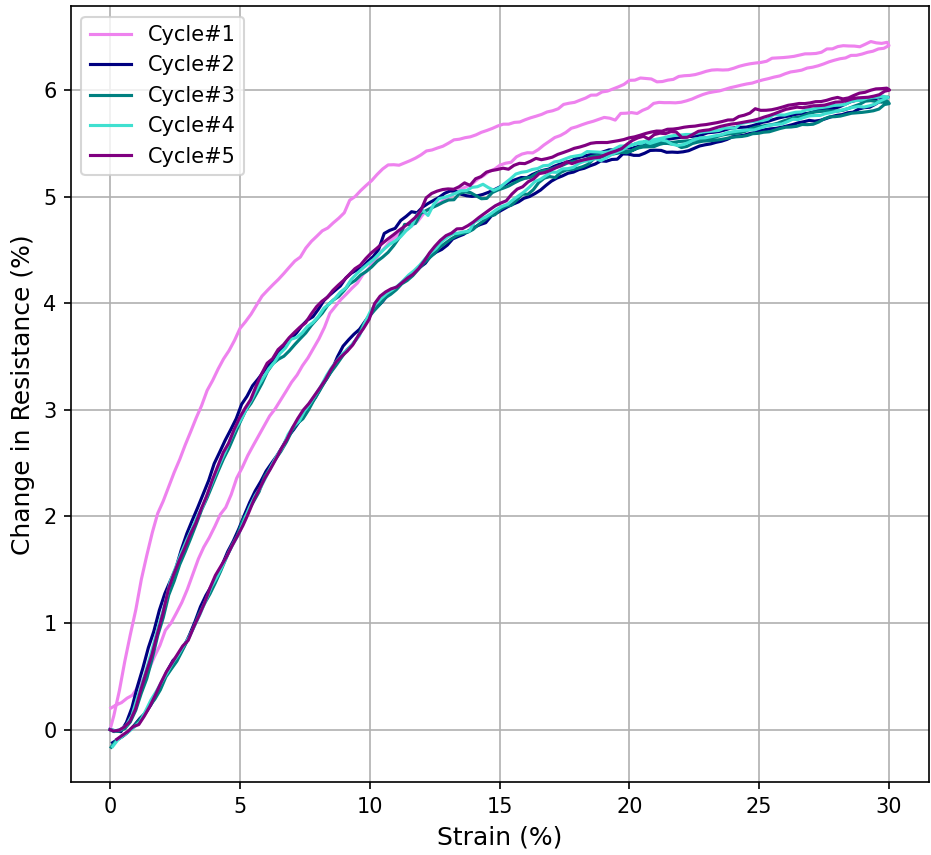}
    \end{minipage}
    \caption{(Left) Images showing the sensor fiber repair process. In the top image, Sil-Poxy is applied with a toothpick and allowed to cure. The bottom image shows a repaired sensor fiber after curing. (Right) 5 stretch-relaxation cycles to 30\% strain with a repaired fiber.}
    \label{fig:repair}
\end{figure}

Cut fibers can be repaired using silicone epoxy (Sil-Poxy; Smooth-On, Inc.) and a small patch of silicone tubing. We threaded the electroplated thread from each cut end through two needles, which were then drawn through the patch of tubing in opposite directions. The patch was then connected to each cut end of the fiber using Sil-Poxy. Because the patched area gains length without gaining electroplated thread, there is a reduction in sensitivity (Figure \ref{fig:repair}). %
There is also an increase in sensor noise across the patched area, which may be due inconsistencies in the repair. %
The repaired sensor stabilizes after the first stretch-relaxation cycle, perhaps due to wear-in of the material. %

\subsection{Effect of long-term wear on resistive sensing}

We stretched one sensor fiber 1000 times to examine the long-term behavior (Figure \ref{fig:baked_long}). %
Resistance readings drifted slowly over the course of the experiment. %
We conclude from this that when using resistance readings to measure force or strain, it is important to calculate a change in resistance relative to an initial reading taken at rest rather than estimating strain or force from absolute resistance measurements. %

\subsection{Degradation in adverse temperature conditions}
\begin{figure}
    \centering
    {\tiny 1000 stretch-relaxation cycles \hspace{70px} Baked sensor fibers}
    \includegraphics[height=90px]{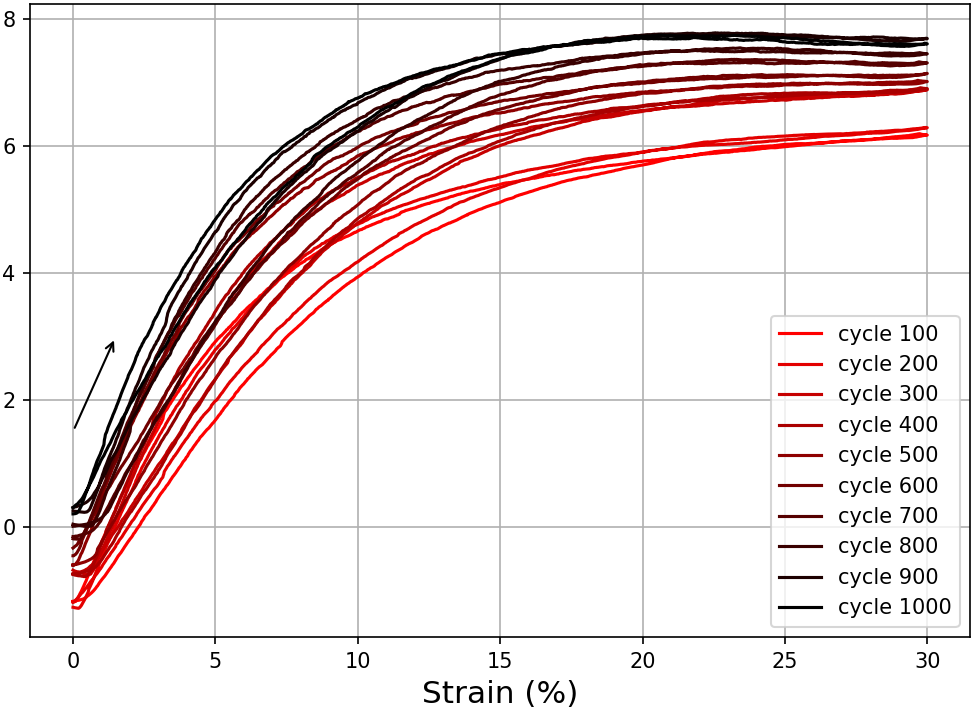}
    \includegraphics[height=90px]{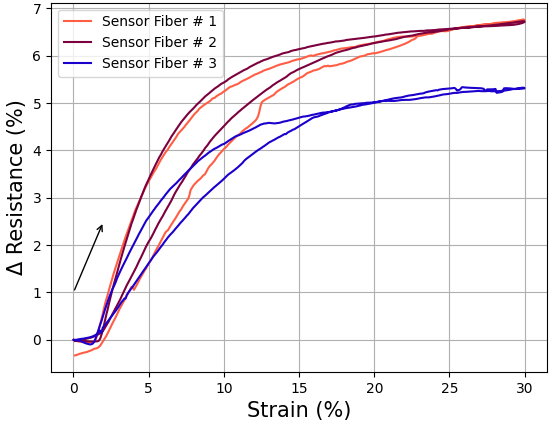}

    \caption{Degradation of sensor fiber under extreme conditions. The arrow indicates that the start of a cycle is on the top left side. (Left) Percent change in resistance for every 100th stretch-relaxation cycle as one fiber is stretched 1000 times. (Right) Relationship between strain and percent change in resistance for three sensor fibers baked for one hour at 55 C. }
    \label{fig:baked_long}
\end{figure}

We baked three sensors at 55 C for one hour, allowed them to cool to room temperature, and performed five stretch-relaxation cycles to 30\% strain using the same experimental protocol as in Section \ref{sec:res_exp}). %
We compare the last stretch-relaxation cycle for each of the three sensors in Figure \ref{fig:baked_long} to show variation across sensors. %
The sensitivity of the baked sensors was degraded but they were still broadly functional. %

\section{Strain sensing applications for robotics}
\label{sec:resistive_applications}

In this section, we demonstrate how the fibers can be used for resistive strain sensing. %
First, we demonstrate a trigger for an assistive grasping glove. %
Second, we show that the fibers are consistent and sensitive enough to be used for proportional-derivative control with only a simple linear fit with no hysteresis modeling. %
Finally, we show how hysteresis and non-linearity can be addressed with a long short-term model using a lookback window of one second. %

\def \armpicheight {32px}
\begin{figure*}
\centering
\vspace{5px}
\begin{minipage}{0.13\textwidth}
\includegraphics[height=\armpicheight]{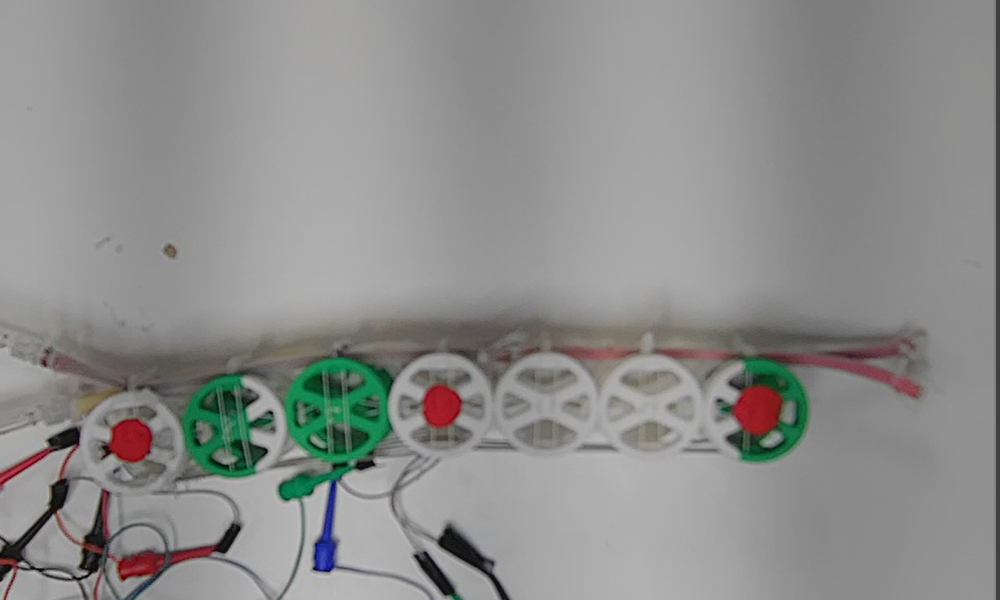}
\\[2px]
\includegraphics[height=\armpicheight]{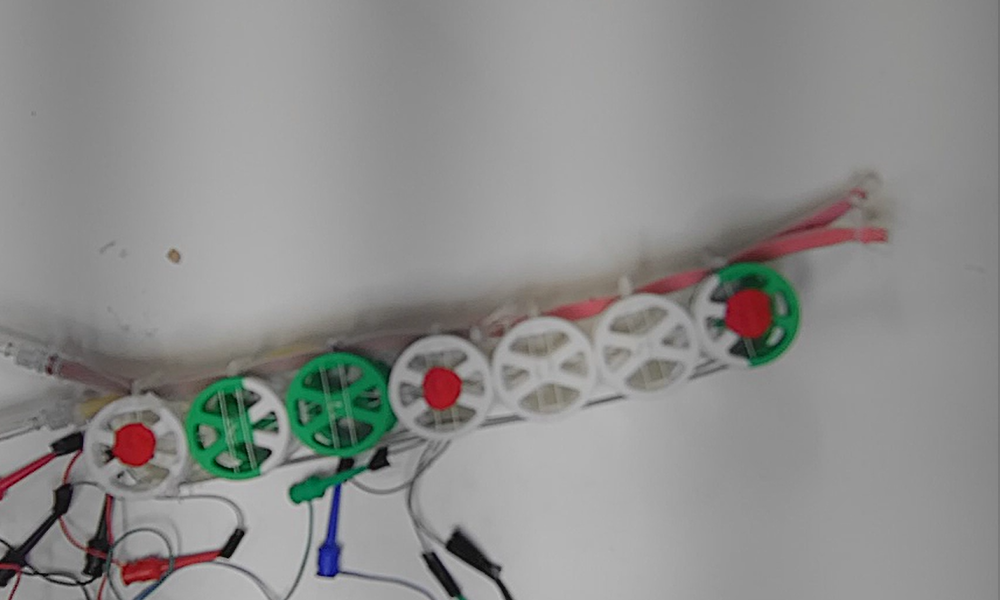}
\\[2px]
\includegraphics[height=\armpicheight]{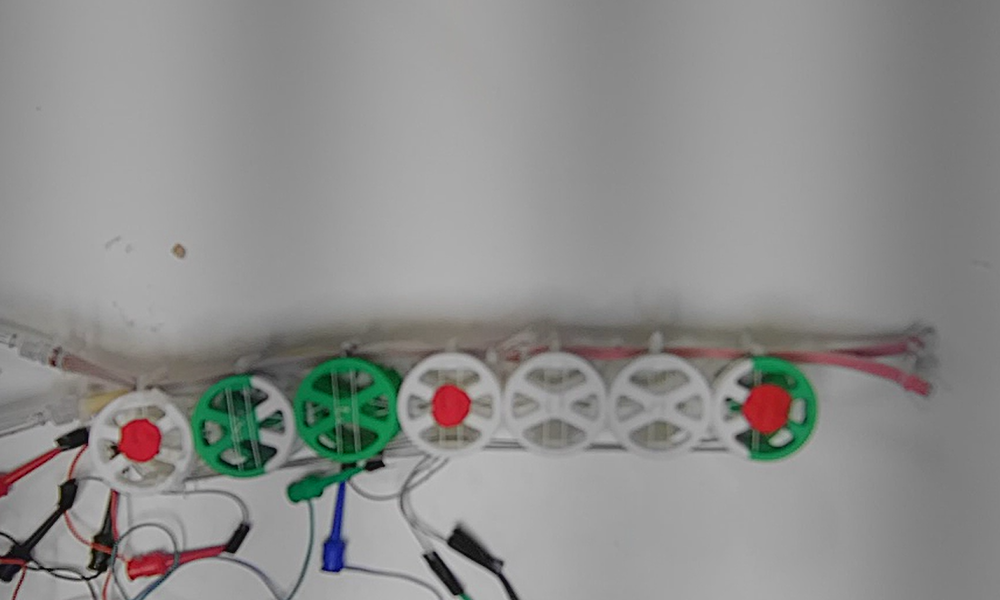}
\end{minipage}
\begin{minipage}{0.72\textwidth}
\includegraphics[width=\textwidth]{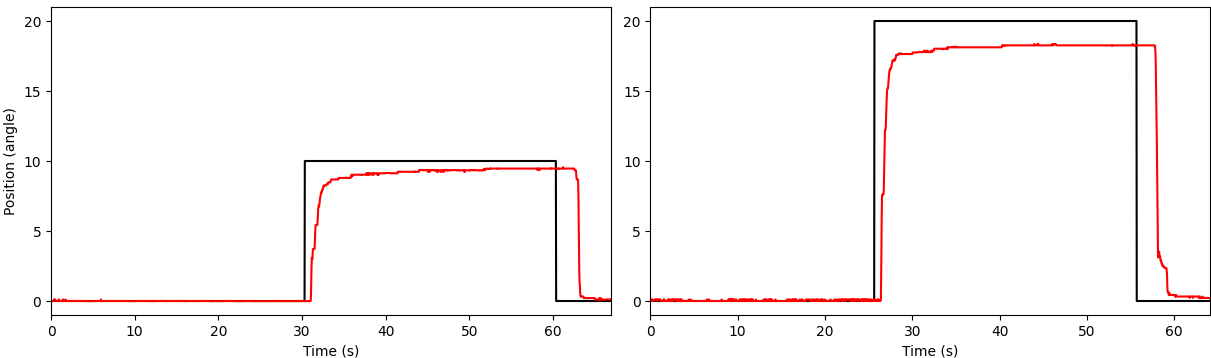}
\end{minipage}
\begin{minipage}{0.13\textwidth}
\raggedleft
\includegraphics[height=\armpicheight]{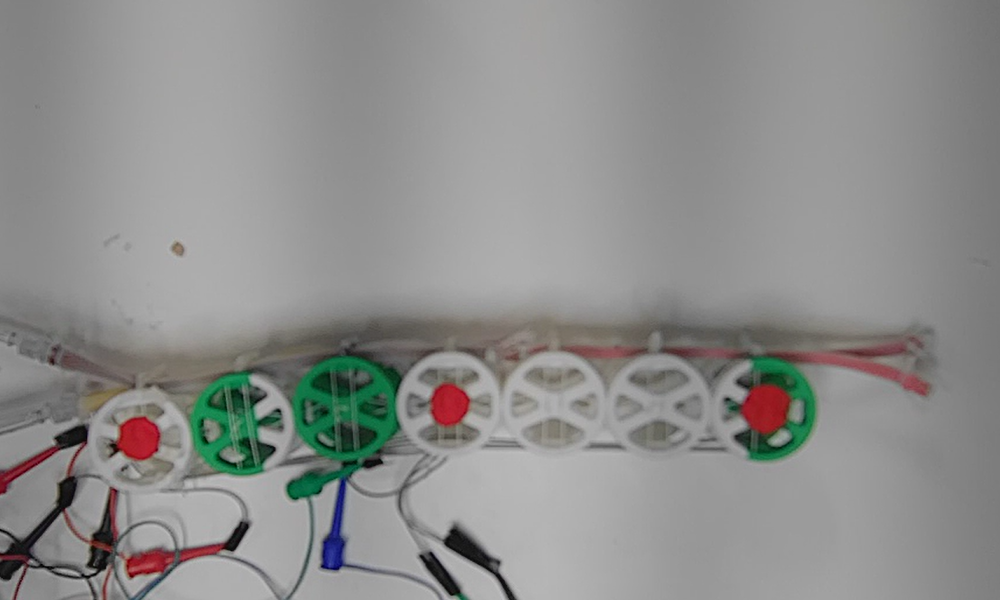}
\\[2px]
\includegraphics[height=\armpicheight]{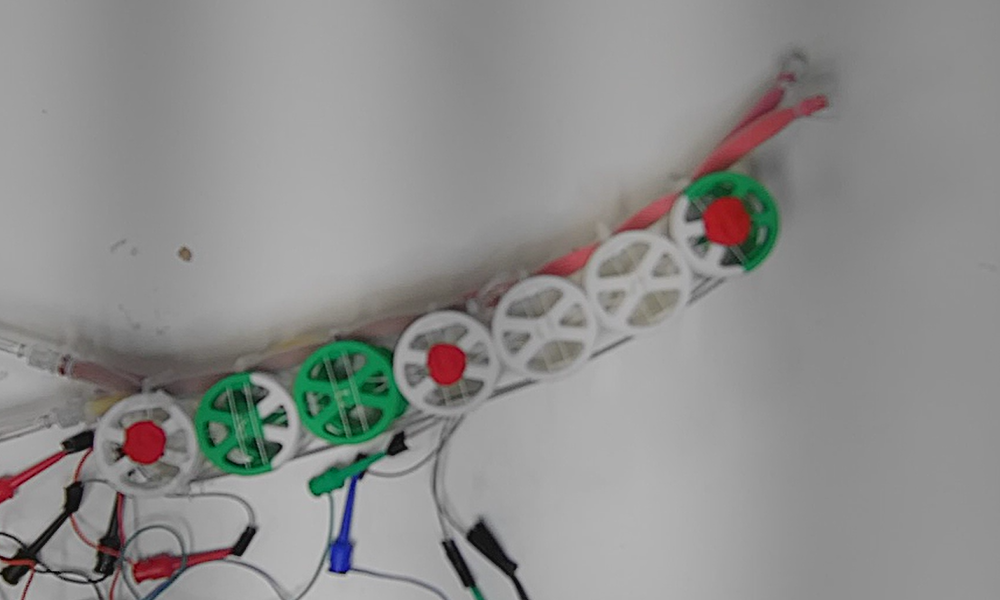}
\\[2px]
\includegraphics[height=\armpicheight]{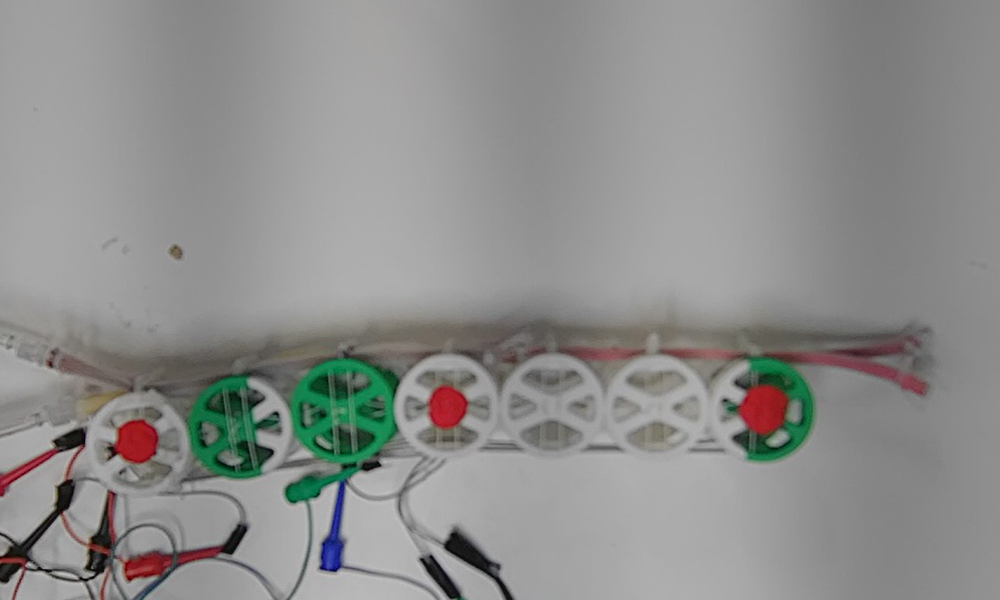}
\end{minipage}
\caption{A comparison of the commanded and actual positions of the robotic strap, with video stills of the robotic strap moving. In both plots, black is the commanded position and red is the actual position as measured using our custom built computer vision based pose detection system. (Left) Data for 10 degree comamnd. (Right) Data for 20 degree command.}
    \label{fig:strap_position_control}
\end{figure*}

\begin{figure}
    \centering
    \includegraphics[width=\linewidth]{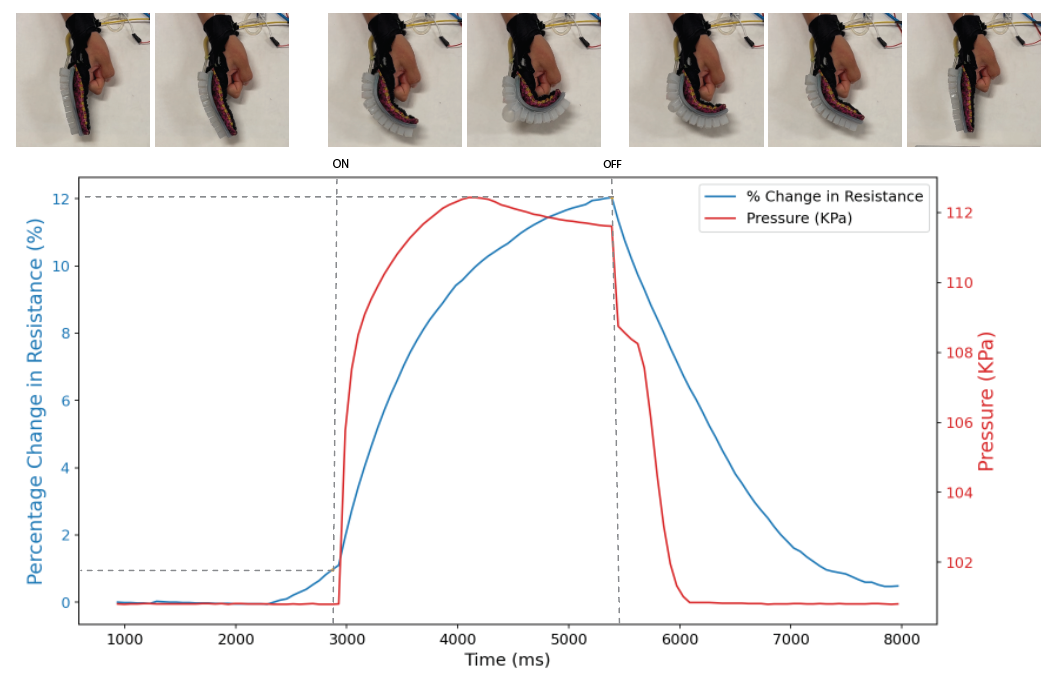}
    \caption{We used a simple threshold to detect stretch as the user's finger was bent. The threshold triggered an assistive grasping behavior.}
    \label{fig:finger}
\end{figure}

\subsection{Grasp triggering in an assistive glove}

\subsubsection{Physical design of assistive glove}
We built a one-finger version of the assistive glove in \cite{polygerinos2013towards} for individuals with partial hand mobility. The glove is made from elastic fabric, with sewn guides for the sensor fiber along the top layer. The strain sensor was inserted through these guides. We sewed an additional elastic fabric layer over the top, securing the sensor between the two layers. 

After preparing the fabric base and integrating the strain sensor, we fabricated the PneuNet pneumatic actuator \cite{polygerinos2013towards} from Ecoflex 00-30 Silicone. %
When air is pumped into the actuator, a series of chambers on one side fill with air and expand while a strain-limiting layer (woven fabric) embedded into the other side prevents expansion, causing the actuator to bend. %
We sewed the PneuNet onto the top fabric layer to integrate it into the glove design.

\subsubsection{Pneumatic control of assistive glove}
We inserted a silicone tube into one end of the PneuNet and connected it to an outlet port of a 3-way valve mini solenoid (Adafruit, part number FA0520E). %
The inlet port was attached to an air compressor (California Air Tools, 8010), and the additional outlet port was attached to the pressure sensor (Adafruit MPRLS), respectively. %
The control board was constructed using a voltage divider circuit to convert sensor output resistance to voltage, two transistors (IRF520), two flyback diodes (1N4001), an Arduino Mega 2560, and a pressure sensor. %
The Arduino continuously reads the sensor's output. %

\subsubsection{Grasp triggering using strain sensor fibers}
When the user begins to curl their finger, the Arduino switches the transistors, triggering the solenoid to activate. %
This causes the PneuNet to inflate and bend, assisting the user's grasp. %
Changes in resistance and pressure inside the actuator over time are shown in Figure \ref{fig:finger}. %
See our \href{https://github.com/Wesleyan-Soft-Robots-Lab/ArduinoMotors}{GitHub repository}. %

\subsection{Position control for a robotic strap}
\label{sec:tentacle}

\subsubsection{Physical design of robotic strap}
We built a pneumatically actuated robot using the physical design of an assistive strap \cite{barhydt2023high}. %
Our robot consisted of seven joints 3D printed out of PLA and tied together using cotton crochet thread with McKibben actuators along one side, such that the strap bends when the actuators are activated. %
The materials for the McKibben actuators were long balloons (Amazon), 
1/4'' expandable sleeving (McMaster, PN 9284K612), barbed tube fittings (McMaster PN 5463K73), and cable ties (McMaster PN 6614K31). Each set of McKibben actuators controls bending of four joints (middle joint shared between bending zones), creating two bending zones. %

Our joints are 20\% larger than in \cite{barhydt2023high} and we mirrored the grooves on both sides to provide a guide for the stretch sensor fibers. %
We attached our sensor fibers to the opposite side of the strap from the Mckibben actuators using zipties, using the middle two for sensing the upper four joints, and the outside two for sensing the lower four joints (Figure \ref{fig:strap_position_control}).

\subsubsection{Pneumatic control board for robotic strap}
We used an Arduino Mega 2560 to control the airflow from a pressurized tank (California Air Tools, 8010) to an air divider mesh. %
Each McKibben actuator was connected to one solenoid (Adafruit, 4663), and each of the two bending zones contained two McKibben actuators actuated in parallel. In manual control mode, the bending was controlled by one dial potentiometer per bending zone (Adafruit, 356). 
 See our \href{https://github.com/Wesleyan-Soft-Robots-Lab/ArduinoMotors}{GitHub repository}. 
 
\subsubsection{Position control using strain sensor fibers}
Two sensor fibers were attached to each of the two joints. %
We manually moved the strap to five locations, and recorded the resistance readings at each location. %
Video data taken from above provided ground truth for the overall position of the arm for each of these readings. %
We used the OpenCV library to identify the centroids of three red spots on the arm, and calculated the angle between the tip of the arm as it bent and the vertical axis. %
The location of the vertical axis was determined by the bottom joint, which was glued down. %

We fitted a linear model to predict the overall curvature from the percent change in resistance. %
We programmed the Arduino to use proportional control on the position estimates to command the robotic strap to move to a desired angle, hold the position for 30 seconds, and return to its initial position. %
See Figure \ref{fig:strap_position_control} for a comparison of the robot's commanded to actual positions for two angle commands. %

\subsection{Improving accuracy with machine learning}
\label{sec:machine_learning}
The two previous demonstrations of resistive strain sensing use only linear correlations between resistance and displacement, showing what can be achieved using only the low-cost, low-technology tools which might be available to a DIY maker. %
For better resourced makers, it is possible to achieve significantly greater accuracy by modeling the hysteresis and non-linearity in the sensors (Figure \ref{fig:stretch_cycles_short}, Section \ref{sec:resistive_experiments}). %

We adapted the technique described in \cite{albaugh2019digital} to knit a sensor sleeve with five vertically oriented sensor fibers on a Brother 891 tabletop knitting machine. %
We fitted this sleeve onto a long, thin latex party balloon (Amazon). %
One end of the balloon was fixed to the table, and the other end was attached to a string which allowed the balloon to be moved by hand. %
We generated a dataset by manually bending the sleeved balloon while taking ground truth position data from an overhead camera. %
We trained a long short-term memory (LSTM) model with two hidden layers of 16 nodes each on this data set. 
The LSTM achieved $\sim$95\% test accuracy with a lookback window of 40 samples (= 1 second). %
We deployed the trained model in real time on an Arduino Mega 2560 (Figure \ref{fig:machine_learning}). %
The estimated position was very accurate in the low-to-medium bending regime (top image), and sometimes lagged slightly behind the ground truth position when the balloon was bent to its maximum bending angle (bottom image). %
See our \href{https://github.com/Wesleyan-Soft-Robots-Lab/ArduinoMotors}{GitHub repository} for more details.

\def \lstmheight {73px}
\begin{figure}
    \centering
    \vspace{10px}
    \begin{minipage}{0.69\linewidth}
    \centering
    {\tiny \qquad Test set accuracy with increasing lookback window}
    \includegraphics[width=\linewidth]{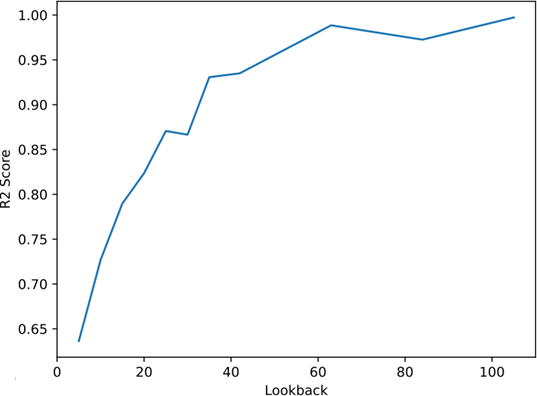}
    \end{minipage}
    \begin{minipage}{0.29\linewidth}
    \raggedleft
    \includegraphics[height=\lstmheight]{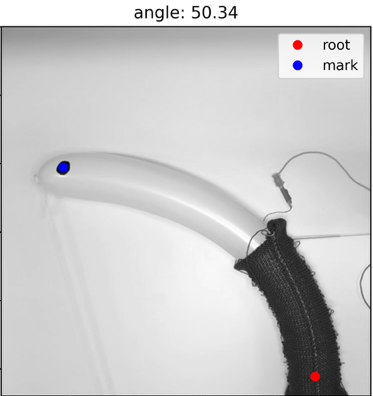}
    \includegraphics[height=\lstmheight]{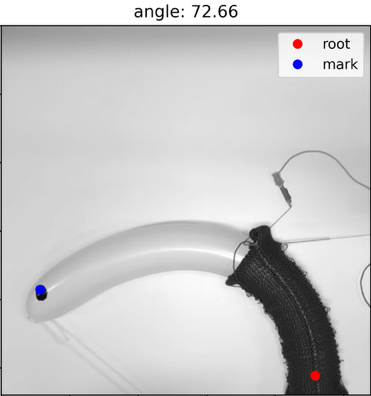}
    \end{minipage}
    \caption{Test set accuracy with increasing lookback window size, and two example still images from the LSTM running on the Arduino annotated with the detected location. There is a black circle on the balloon itself which was used for ground truth tracking by an overhead camera. The blue dot labeled ``mark" shows the position estimated by the LSTM, while ``root'' is the base of the balloon, used for calculating the angle of bending. }
    \label{fig:machine_learning}
\end{figure}

\section{Capacitive sensing robotics applications}
\label{sec:capacitive_applications}

In this section, we demonstrate how the fibers can be used for capacitive sensing. %
Because silicone tubing is extremely flexible and does not kink, the sensor fibers can be knitted. %
We created two kinds of sensors by knitting squares of fiber: a touch sensor and a near-field sensor. %
For both sensors, we used the capacitance sensing circuit from \cite{xu2024cushsense}. %

\subsection{Capacitive sensor integration with a robot arm}

To integrate a capacitive touch sensor with a UFACTORY xArm 6 DoF robotic arm, we used an Arduino Mega 2560 to receive capacitance readings from the capacitance sensing circuit \cite{xu2024cushsense} and send them to a desktop computer. %
A Python script run on the desktop computer used PySerial to read the sensed capacitance from the Arduino and the xArm library to send commands to the arm. %
See our \href{https://github.com/Wesleyan-Soft-Robots-Lab/Capacitive-Sensing-Sleeve}{GitHub repository}. %

\subsection{Touch sensor for collision avoidance}
\label{sec:touch}
\begin{figure}
    \centering
    \vspace{10px}
    \hspace{4mm}
    \includegraphics[width=0.3\linewidth]{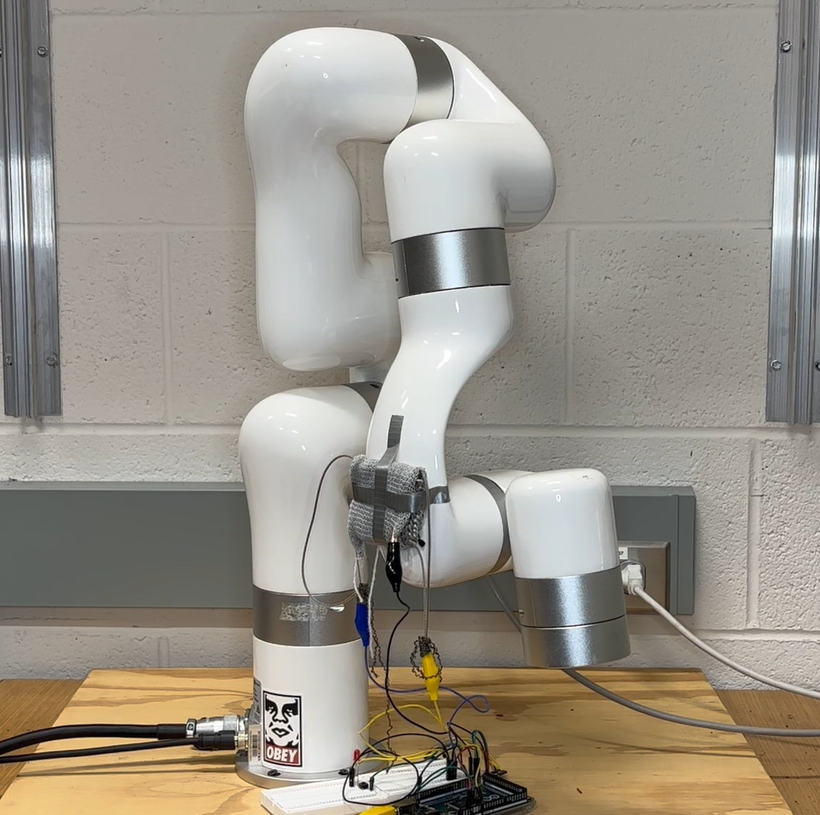}
    \includegraphics[width=0.3\linewidth]{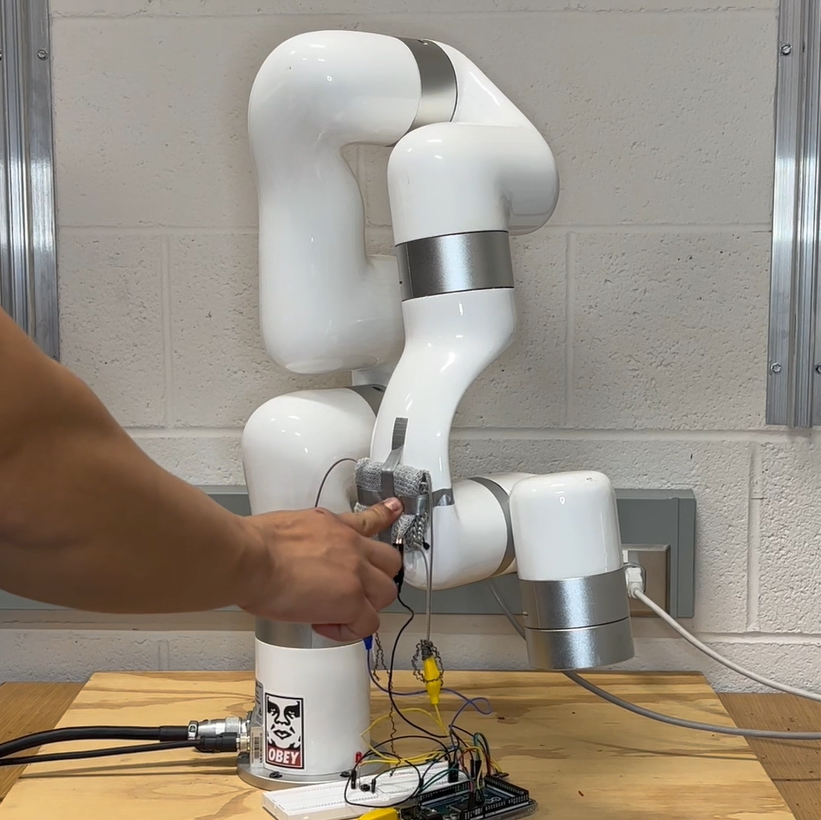}
    \includegraphics[width=0.3\linewidth]{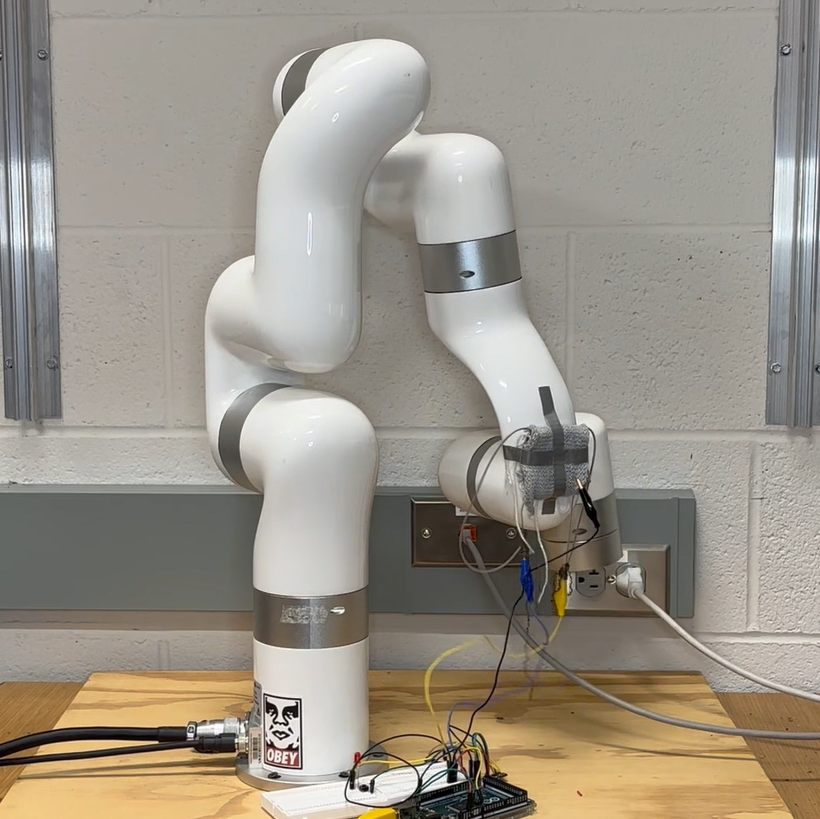}
    \\\vspace{1mm}
    \includegraphics[width=\linewidth]{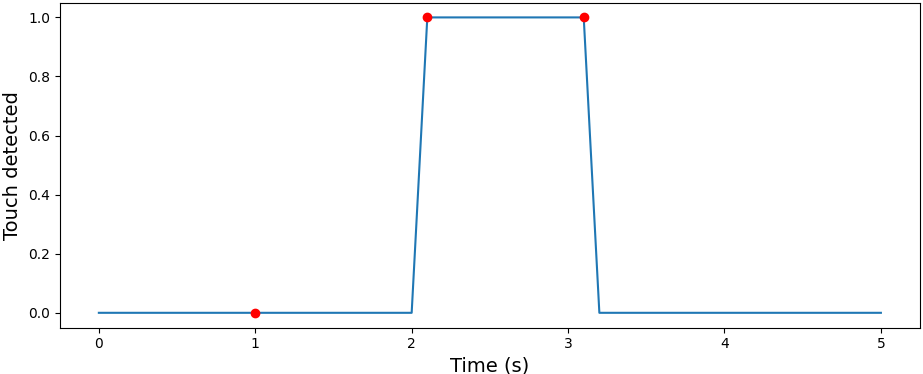}
    \caption{We used a simple binary classifier to detect touch (Section \ref{sec:touch}). The red dots show the state of the capacitive sensor in the three images above. Our initial state is ``no touch.'' As the robot is touched, the capacitance threshold rises and the robot's mode switches to ``touch,'' triggering the robot arm to move away from the direction of the touch. }
    \label{fig:touch_sensor}
\end{figure}

We hand knitted two 10' sensor fibers into rectangular patches using a plain knit stitch (stockinette), which we then sewed together into a pair of ``electrodes'' using acrylic thread. %
We used the plain knit stitch for this application to increase the surface area of the touch sensor. %
The plain knit stitch produces a larger surface area for a given length of fiber than garter stitch, and because we needed to sew two squares together, we could prevent the two squares from curling. %
To provide passive shielding, we wrapped the electrode pair in fabric knitted from composite conductive thread (80\% acrylic, 20\% stainless steel blend, purchased from Yeoman Yarns, Leicester, UK). %
The touch sensor was then taped to the robot arm, and the robot was programmed to move away from the location of the patch when a certain threshold of capacitance was reached (Figure \ref{fig:touch_sensor}, supplemental video).

\subsection{Near-field sensor for human hand tracking}
\label{sec:near}
We re-knitted the two 10' sensor fibers into rectangles using garter stitch so that the knitted squares would lay flat, as in the characterization experiments. %
In the touch sensing application, we could sew two knitted patches to each other, which would flatten them and prevent curling. %
In this near-field sensing application, we needed the sensor patches to lie flat on their own without sewing two of them together. %

\begin{figure}
    \centering
    \includegraphics[width=\linewidth]{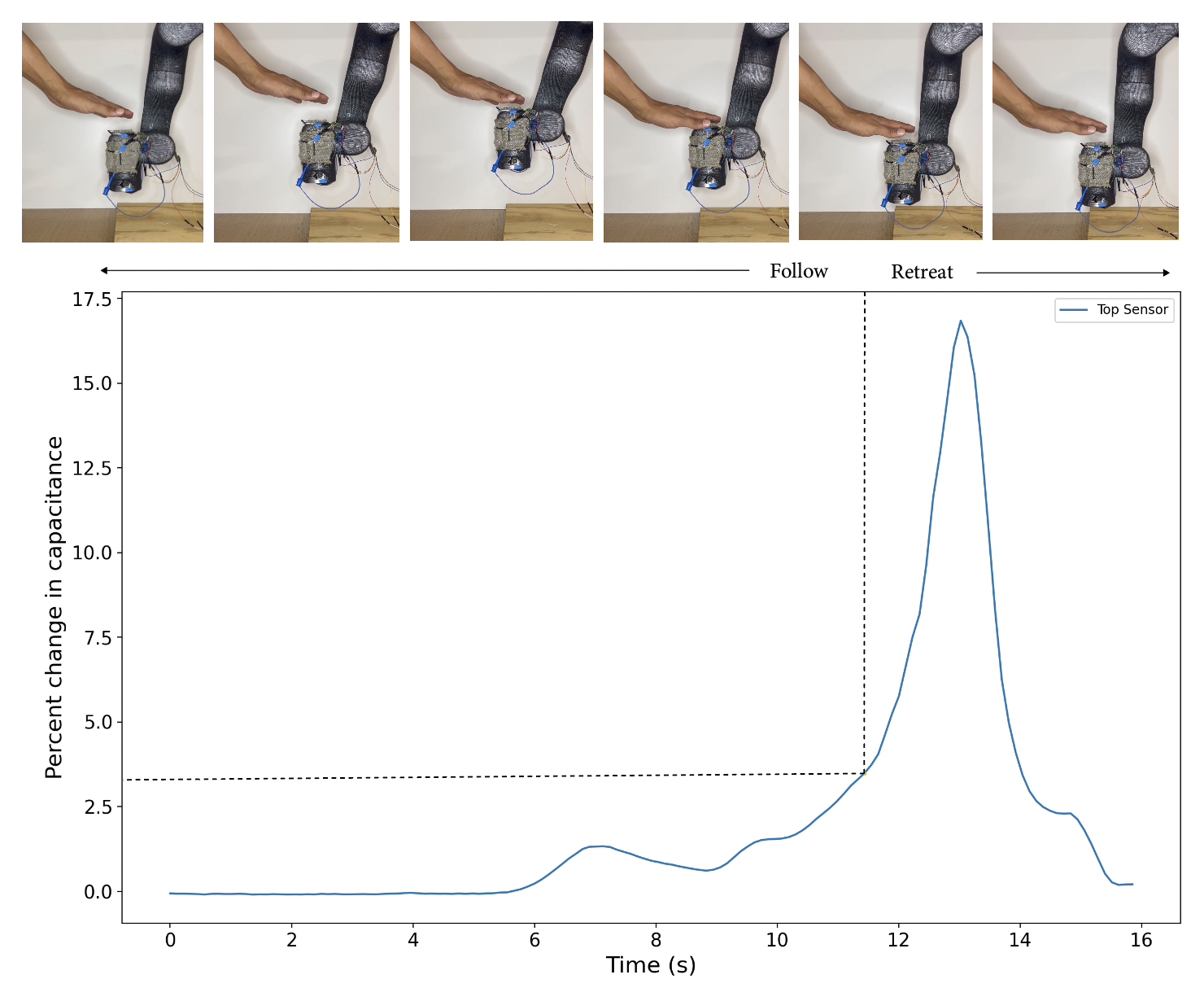}
    \caption{The robot arm was programmed to seek a 3.5\% percent change in capacitance from baseline. At lower sensed capacitance values, the arm would move up, towards the location of the sensor on its body; at higher values, it moved down and away. The robot was thus able to follow the motion of a human hand. See the supplemental video for a demonstration of the arm using two sensors to follow a human hand through space. }
    \label{fig:near_field}
\end{figure}

To use the capacitive sensors for near-field sensing, we considered the hand the ``ground'' electrode and did not use shielding. %
We programmed the robot to maintain set distance from a human hand using PD control. %
When the robot arm could detect the hand but the capacitance reading was below its set point, it moved towards the hand. %
If the hand moved closer to the sensor, the robot would move way until the sensed capacitance level lowered to the set point. %
The robot arm was thus able to track and follow a hand moving through space (Figure \ref{fig:near_field}). %
We chose 3.5\% change from baseline as the set point for this example because the distance is towards the limit of the robot's sensing range, and represents a comfortable --- if short --- distance for human-robot interaction. %
See the supplemental video for a demonstration with two sensors used simultaneously. %

\section{Adaptations for scalable manufacturing}
\label{sec:scalability}
While the manufacturing process only takes a few minutes for the short lengths of fiber used in most of the experiments described here, making long lengths of fiber --- needed for applications like the touch-sensitive patches in Section \ref{sec:capacitive_applications} --- can be time consuming. %
It is also much easier to knit with an industrial machine if the fiber can be drawn steadily from a spool. %
We discuss two adaptations to increase scalability. %

\subsection{Under-stuffing the fibers} 
Because the capacitive touch and near-field sensitive knitted patches do not require extreme deformation, large quantities of fiber for these applications can be made by loosely rahter than densely stuffing the tubing with electroplated thread. %
This can reduce the manufacturing time more by more than half, and still requires no specialized equipment or additional tools. %
It also preserves certain other desirable properties such as repairability and the hypoallergenic properties associated with silicone. %
This manufacturing process represents a plausible middle ground for a maker who needs to create a few meters of fiber, but does not need to create hundreds of meters of fiber. %
However, this manufacturing method still requires human attention and is not scalable to industrial production. %
This method also will only work for the capacitive applications: Strain sensing applications still require larger deformations and therefore fully stuffed tubing. %

\subsection{Filling the tubing with conductive fluid}
\label{sec:conductive_fluid}
A truly scalable, yet still relatively low cost, version of these sensor fibers can be made by injecting conductive fluids through tubing. %
This method was inspired by soft sensors using liquid metal and ionic solutions, such as \cite{chossat2013soft}, but reduces manufacturing cost by using off-the-shelf parts and only inexpensive conductive fluids rather than liquid metal. %
The difficulty with using a fluid as the conductive element in a sensor like this is that the fluid must not be able to leak or evaporate out of the tubing. %
In order to investigate the viability of this sensor manufacturing method, we performed a proof-of-concept experiment with a small number of fluid-filled fibers, and tracked their change in resistance over time. %

We tested two tubing materials: Silicone and latex. %
Silicone tubing is more ubiquitously available, but is permeable to water vapor due to having larger pores than latex. %
We made one example fiber using each of four low cost and relatively non-toxic conductive solutions: 0.111 g CaCl$_2$/mL L in a 1:1 (by volume) water/ethylene glycol mixture, 0.111 g CaCl$_2$/mL in water, 0.58 NaCl/mL in water, and a saturated NaCl solution in ethylene glycol. %
Ethylene glycol was selected due to its low viscosity and low volatility. %

\begin{figure}
    \centering
    \vspace{10px}
    \includegraphics[width=0.9\linewidth]{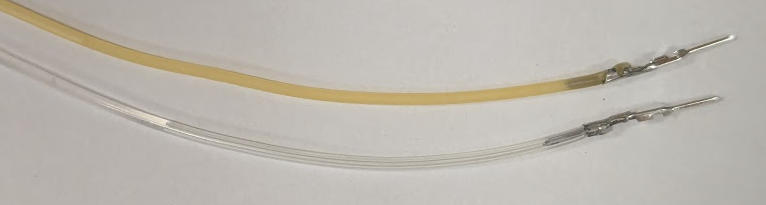}
    \caption{Two sensor fibers made with a scalable manufacturing method using conductive fluids injected through latex and silicone tubing (Section \ref{sec:conductive_fluid}). Silicone is porous to water vapor, and a large air bubble can be seen in the silicone tube (bottom). }
    \label{fig:latex_silicone_sensors}
\end{figure}

We manufactured the conductive fluid-filled sensor fibers by injecting the fluid through a 20 cm length of tubing using a syringe, placing a short length (1 cm) of stainless steel wire into the tubing, and then crimping the tubing over the stainless steel wire (Figure \ref{fig:latex_silicone_sensors}). %
We measured the resistance at rest for each fiber 7 times over the next 14 days. %

All of the silicone tubing fibers lost conductivity within 7 days. %
The latex fiber with the CaCl$_2$ water-based solution experienced an order of magnitude increase in resistance over the 14 days (initial value 0.8 M$\Omega$; ending value 7.5 M$\Omega$), and the latex fiber with the NaCl water-based solution steadily increased before eventually losing connectivity on the 14th day (initial value 0.7 M$\Omega$; last measured value 45 M$\Omega$). %

The latex fibers with the ethylene glycol-based conductive solutions showed more promise. %
After an initial reading of 2 M$\Omega$, the fiber with the CaCl$_2$ ethylene glycol/water-based solution displayed a consistent reading of 6 M$\Omega$ for the rest of the 14 day period. %
The fiber with the NaCl ethylene glycol-based solution also stayed conductive, with an initial reading of 3 M$\Omega$ and a last reading of 5 M$\Omega$, but the intermittent readings varied more (ranging from 2 M$\Omega$ to 3.5 M$\Omega$). %
We will perform a more detailed sweep of conductive solutions and tubing materials in future work. %

While this manufacturing method does create a scalable version of the fibers using low cost materials, there are trade-offs with the cost of the equipment and the repairability of the fibers. %
Laboratory supplies, mixing equipment, crimping tools, and pumps to fill larger lengths of tubing represent a significant increase in equipment costs compared with a \$2 needle. %
Also, whereas the electroplated thread fiber can be easily repaired (Section \ref{sec:repair}, supplemental video), if the conductive fluid leaks out of a sensor fiber, the tubing would need to be patched and waterproofed. %
Then both crimped ends would need to be removed, the fluid reinjected, and the crimps replaced. %
This version of the sensor fibers is therefore more appropriate for use in a research lab or by an industrial manufacturer than by a hobbyist, educator, or DIY maker. %

\section{Conclusions} 
\label{sec:conclusion}
We present a low-cost method for manufacturing resistive and capacitive sensor fibers. %
Our manufacturing method requires no specialized equipment and uses only easily accessible and inexpensive off-the-shelf parts, making it perfectly suited for DIY, community, and educational settings. %
For researchers and makers with greater access to resources, we also present options to improve accuracy (Section \ref{sec:machine_learning}) and scale production (Section \ref{sec:scalability}), and discuss their trade-offs. 

The shape and flexibility of these sensor fibers makes them easy to incorporate in a wide variety of applications (Sections \ref{sec:resistive_applications}, \ref{sec:capacitive_applications}). %
They can be stretched along their length and used for resistive strain sensing (Section \ref{sec:resistive_experiments}) or used for capacitive touch and near-field sensing (Section \ref{sec:capacitive_applications}). %
Also, because of their extreme flexibility, they can even be knitted or knotted (Section \ref{sec:capacitive_experiments}). %
They are durable (Section \ref{sec:repair}) and, because they are repairable, they can also be re-used for applications requiring different lengths of material. %

\addtolength{\textheight}{-12cm}   




\section*{ACKNOWLEDGMENT}
This work was supported by NSF ERI award \#347708. 
We thank Caryn Canalia for administrative support and Ben Parker for machining support.


\bibliographystyle{ieeetr}

\bibliography{references}

\end{document}